\documentclass[10pt,twocolumn,letterpaper]{article}

\usepackage{cvpr}              
\usepackage{multirow}
\usepackage{graphicx}
\usepackage{soul}
\usepackage{subcaption} 
\usepackage[accsupp]{axessibility}

\usepackage[dvipsnames]{xcolor}

\definecolor{cvprblue}{rgb}{0.21,0.49,0.74}
\usepackage[pagebackref,breaklinks,colorlinks,citecolor=cvprblue]{hyperref}

\def\paperID{102\_69} 
\def\confName{CVPR}
\def\confYear{2024}

\title{Evaluating the Effectiveness of Video Anomaly Detection in the Wild\\ 
\emph{Online Learning and Inference for Real-world Deployment}}

\author{Shanle Yao\\
{University of North Carolina Charlotte}\\
{\tt\small syao@charlotte.edu}
\and
Ghazal Alinezhad~Noghre\\
{University of North Carolina Charlotte}\\
{\tt\small galinezh@charlotte.edu}
\and
Armin Danesh~Pazho\\
{University of North Carolina Charlotte}\\
{\tt\small adaneshp@charlotte.edu}
\and
Hamed Tabkhi\\
{University of North Carolina Charlotte}\\
{\tt\small htabkhiv@charlotte.edu}
}

\begin{document}
\maketitle
\begin{abstract}

Video Anomaly Detection (VAD) identifies unusual activities in video streams, a key technology with broad applications ranging from surveillance to healthcare. Tackling VAD in real-life settings poses significant challenges due to the dynamic nature of human actions, environmental variations, and domain shifts. Many research initiatives neglect these complexities, often concentrating on traditional testing methods that fail to account for performance on unseen datasets, creating a gap between theoretical models and their real-world utility. Online learning is a potential strategy to mitigate this issue by allowing models to adapt to new information continuously. This paper assesses how well current VAD algorithms can adjust to real-life conditions through an online learning framework, particularly those based on pose analysis, for their efficiency and privacy advantages. Our proposed framework enables continuous model updates with streaming data from novel environments, thus mirroring actual world challenges and evaluating the models' ability to adapt in real-time while maintaining accuracy. We investigate three state-of-the-art models in this setting, focusing on their adaptability across different domains. Our findings indicate that, even under the most challenging conditions, our online learning approach allows a model to preserve 89.39\% of its original effectiveness compared to its offline-trained counterpart in a specific target domain.


\end{abstract}    
\section{Introduction}
\label{sec:intro}

Video Anomaly Detection (VAD) is an essential area within computer vision, tasked with pinpointing atypical behaviors in specific scenes. It plays a pivotal role in a variety of sectors, including surveillance and healthcare, where identifying deviations from the norm is crucial. The scope of anomalies it addresses is wide, with a significant emphasis on detecting anomalies centered around human activities. VAD techniques are primarily categorized into two types: pixel-based methods, which analyze the raw data of pixels, and pose-based methods, which focus on the dynamics of joints and bodily movements. The latter is especially beneficial in scenarios where privacy is a major concern, as it prioritizes the analysis of skeletal movements over detailed pixel imagery. This approach minimizes privacy concerns and plays a vital role in mitigating biases, particularly those affecting marginalized communities, thus providing a fairer and more privacy-aware anomaly detection method.

\begin{figure}[]
    \centering
    \includegraphics[clip,trim={17 17 17 17},width=1\columnwidth]{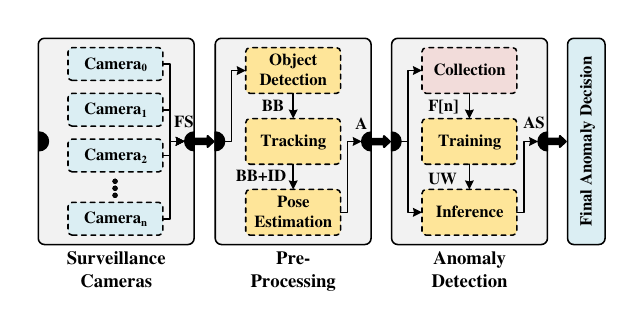}
    \caption{A conceptual overview of an end-to-end system with online unsupervised anomaly detection training. Frame sequences (FS) collected from surveillance cameras pass through a pre-processing phase to extract necessary annotations (A), including bounding boxes (BB), tracking information (ID), and pose information. This information consequently goes through anomaly detection, which is used for real-time inference and collection. The collection algorithm collects enough frame annotations (F[n]) for training. After training, Updated Weights (UW) are replaced for the next inference step.}
    \label{fig:intro}
\end{figure}

Recognizing the wide range of normal and anomalous behaviors, a reflection of human behavior's complex nature poses a significant challenge to the generalizability of VAD models in real-world scenarios. The inability of existing datasets to fully capture this breadth significantly hampers the applicability of VAD models outside laboratory conditions. This limitation has spurred a shift towards unsupervised learning in VAD, where models learn from unlabelled data, recognizing normal behavior patterns and identifying deviations without needing predefined anomalies. This move towards unsupervised methods represents a pivotal adaptation, promising to enhance the robustness and relevance of VAD models in diverse and unpredictable real-world environments by accommodating the full spectrum of human behaviors.

Nonetheless, human-centric VAD faces other inherent challenges, such as the context-specific nature of what constitutes an anomaly. This variability means that behaviors considered normal in one setting might be deemed anomalous in another. For instance, punching in a gym is typical, whereas the same behavior in a mall would be considered anomalous. Domain shift hurdle is more significant when transitioning anomaly detection models from controlled experimental settings to real-world applications of anomaly detection, in which anomalous behaviors are deviations from established norms of behavior. In real-world environments, ostensibly normal behaviors can often be misconstrued as anomalous by these models due to discrepancies arising from various factors, such as camera angles and distance, which were not accounted for during the training phase. Such discrepancies can lead to an inflated rate of false positives, substantially undermining VAD systems' practical utility and accuracy in natural settings. This vulnerability to domain shift underscores the need for more adaptive, context-aware machine learning models capable of dynamically recalibrating their parameters to the nuances of their operational environment, thereby enhancing their effectiveness and reliability in diverse real-world applications.

Existing video anomaly detection (VAD) methodologies often rely on offline learning paradigms, which inherently limit their ability to adapt to real-world situations' dynamic and unpredictable nature. The shift towards online learning for VAD anomaly detection is not merely a trend but a necessary evolution to address these limitations. By continually updating their knowledge base with new, unlabeled data encountered in their operational environment, online learning algorithms embody the adaptability required to tackle the complex nature of human behavior and the broad spectrum of what may be considered abnormal in different contexts. This capability to learn from streaming data in real-time allows for the detection system to remain relevant and practical, even as the nature of anomalies evolves. 

To our knowledge, no existing research has shied away from online learning VAD, specifically within the domain of pose-based VAD, marking a significant gap in the literature. It is also important to separate the concept of "online learning" VAD from the broader concept of "online anomaly detection" as outlined in various studies focused on pixel-based analysis, such as those by \cite{doshi2021online, zhang2023online, karim2024real}, where the term is typically associated with the capacity for real-time decision-making. Unlike mere online anomaly detection, which implies immediate processing without learning from new data, online learning involves the algorithm's ability to continuously adapt and update its understanding, enhancing its predictive accuracy over time. Overall, we observe a notable oversight in mainstream VAD research, where the inherent benefits and necessities of online learning for anomaly detection are often overshadowed by results derived from the offline learning paradigm. 

This study rigorously assesses the effectiveness and adaptability of current pose-based Violence Detection (VAD) methodologies, focusing on their application in online VAD environments that simulate real-life conditions. Our aim is not only to highlight the strengths and weaknesses of each model in the context of online VAD but also to reveal their adaptability and efficiency in transitioning to new domains. To this end, we design and implement an online learning framework that mirrors the actual world challenges of VAD. The proposed online learning VAD framework enables continuous model updates from novel environments, thereby testing their ability to adapt in real-time to new domains while maintaining high levels of accuracy and privacy advantages. We analyze the performance and efficiency of three state-of-the-art models, GEPC\cite{markovitz2020graph}, STG-NF\cite{hirschorn2023normalizing}, and TSGAD\cite{noghre2024exploratory}, in an execution environment that emulates unseen streaming data in the real world. The findings from our experiments demonstrate the proposed online learning frameworks' effectiveness, where models are able to preserve between 89.39\% and 99.20\% of their performance, as measured by the Area Under the Receiver Operating Characteristic Curve (AUC-ROC), in both the worst and best-case scenarios, respectively, relative to models trained offline in the target domain.

In summary, this study presents the following contributions:
\begin{itemize}
    \item Development of an online learning framework tailored for pose-based anomaly detection.
    \item Evaluation of traditional offline learning methodologies to discover their efficacy and limitations within unseen online scenarios.
    \item Highlighting the research gaps and looking at the evolution and potential breakthroughs of video anomaly detection in the wild. 
\end{itemize}
\section{Related Works}
\label{sec:related}

Historically, traditional VAD methods predominantly utilized handcrafted features\cite{cheng2015gaussian, cocsar2016toward, yuan2014online}, which, while effective in controlled settings, often faced challenges in generalizing to the diverse conditions of real-world applications. With the advent of deep learning, a paradigm shift occurred in VAD, leading to its classification into two primary strategies: pixel-based and pose-based approaches. Pixel-based methods\cite{huang2023multi, zaheer2022generative, georgescu2021background, georgescu2021anomaly, barbalau2023ssmtl++, wang2022video, chen2021nm, RL00} analyze the raw pixel data to detect anomalies, whereas pose-based approaches concentrate on extracted skeletal information and monitoring the movements of individuals within the scene. This study specifically focuses on the exploration of pose-based VAD techniques. Consequently, we will provide a detailed examination of pose-based methods in the subsequent sections, highlighting their operational mechanisms and advantages.

In unsupervised learning environments, strategies are developed to establish tasks that inherently encourage models to assimilate normal behavior patterns. These tasks predominantly involve reconstructing the current timestep\cite{yu2023regularity, chen2023multiscale, jain2021posecvae, markovitz2020graph} or predicting future or past sequences\cite{zeng2021hierarchical, huang2022hierarchical, rodrigues2020multi}. Several studies\cite{noghre2024exploratory, Morais_2019_CVPR} employ a multi-branch framework, leveraging both objectives to enhance anomaly detection capabilities, showcasing the diversity and adaptability of unsupervised methods in identifying deviations from established norms.

\begin{figure}[]
    \centering
    \includegraphics[clip,trim={17 17 17 17},width=1\columnwidth]{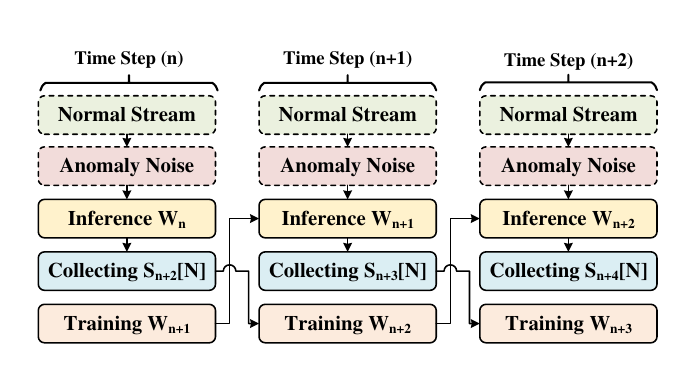}
    \caption{A conceptual overview of an end-to-end system with online unsupervised anomaly detection training.}
    \label{fig:pipeline}
\end{figure}

Among the pose-based models, GEPC\cite{markovitz2020graph}, TSGAD\cite{noghre2024exploratory}, and STG-NF\cite{hirschorn2023normalizing} distinguish themselves by having their source code publicly accessible. Consequently, these models were selected for our experimental analysis. The GEPC model\cite{markovitz2020graph} encodes input pose sequences into a latent graph space, followed by a clustering process. Anomalies are detected through a Dirichlet mixture model that evaluates the distribution of cluster-based action normalities. The STG-NF model\cite{hirschorn2023normalizing} leverages normalizing flows to map input pose sequences into a standard distribution within latent space, with the degree of deviation from this distribution indicating potential anomalies. The TSGAD model\cite{noghre2024exploratory} analyzes anomalies by examining both pose and trajectory data. It employs a graph variational autoencoder for pose analysis, generating scores based on deviations from the model's learned distribution, while the trajectory analysis predicts future movements, comparing these predictions to actual trajectories to produce a trajectory-based anomaly score. The overall anomaly detection is then determined by combining the scores from both the pose and trajectory analyses.

Multiple pixel-based investigations\cite{doshi2021online, zhang2023online, karim2024real} interpret "online anomaly detection" differently from our discussion in \cref{sec:intro}. They consider it as the ability of a model to dynamically render decisions in real time. Contrarily, our manuscript delves into online learning for anomaly detection, highlighting the model's continuous adjustment and training with streaming data to enhance detection accuracy. Furthermore, while several studies\cite{doshi2023towards, danesh2023chad} explore cross-domain evaluation under a zero-shot framework, they fall short of suggesting any strategies for domain adaptation, particularly concerning streaming data.
\section{Methodology}
\label{sec:methodology}

To explore the viablilty of the online unsupervised anomaly detection training application with frameworks using existing pose-based algorithms, a three-stage pipeline was developed, emulating real-world scenarios. This pipelin design illustrated in \cref{fig:pipeline} comprises an inference stage, a collection stage, and a training stage. At the inference stage, algotithm with source pretrained weight is used to process input stream to identify normal behavioral patterns with potential existing anomaly noise. Subsequently, sequences detected as 'normal' are formatted and collected for training purposes within the collection stage. Once the training data is accumulated to pre-defined volume, the training stage will start  fine-tuning the pre-trained weights to adapt to the target domain weight with evaluation. Notably, the inference stage's weights are updated with these refined weights with a lag of two time steps from the initial state because of the nature of such a pipeline design.

\subsection{Inference Methodology}

\begin{table}[]
\centering
\caption{Number of Poses comparison for ShanghaiTech\cite{shanghaitech}, CHAD\cite{danesh2023chad}, and different cameras views from CHAD \cite{danesh2023chad}}
\label{tab:dataset}
\begin{tabular}{c|ccc}
\multirow{2}{*}{\textbf{Dataset}} & \multicolumn{3}{c}{\textbf{Number of Poses}} \\ \cline{2-4} 
 & \multicolumn{1}{c|}{\textbf{Train}} & \multicolumn{1}{c|}{\textbf{Test}} & \textbf{Total} \\ \hline
ShanghaiTech\cite{shanghaitech} & \multicolumn{1}{c|}{257,650} & \multicolumn{1}{c|}{37,845} & 295,495 \\ \hline
CHAD\cite{danesh2023chad} & \multicolumn{1}{c|}{802,167} & \multicolumn{1}{c|}{119,867} & 922,034 \\ \hline
CHAD\cite{danesh2023chad} Cam 0 & \multicolumn{1}{c|}{111,230} & \multicolumn{1}{c|}{21,074} & 132,304 \\ \hline
CHAD\cite{danesh2023chad} Cam 1 & \multicolumn{1}{c|}{213,991} & \multicolumn{1}{c|}{35,502} & 249,493 \\ \hline
CHAD\cite{danesh2023chad} Cam 2 & \multicolumn{1}{c|}{245,436} & \multicolumn{1}{c|}{35,727} & 281,163 \\ \hline
CHAD\cite{danesh2023chad} Cam 3 & \multicolumn{1}{c|}{231,510} & \multicolumn{1}{c|}{27,564} & 259,074
\end{tabular}%

\end{table}

\subsubsection{Input Stream}
As shown and discussed in \cref{fig:intro} from \cref{sec:intro}, prerequisite for real-world applications of online anomaly detection is the capability to accurately detect and track individual figures within video streams to extract sequential pose information. This process could be easily influenced by noise, primarily due to the variability introduced by diverse streaming conditions such as location, camera angles, and coverage area.These factors contribute to the domain-specific nature of anomaly detection tasks. For instance, jogging or running, which is normal behavior in a park setting, may be considered anomalous within a grocery store environment.

To minimize the impact of such variability and enhance the precision of extracted pose sequences, pose-based datasets are employed to emulate real-world streaming conditions. Among existing pose based anomaly datasets with continues pose sequences, CHAD dataset\cite{danesh2023chad} offers a comprehensive collection of 922,034 count of pose instances captured from four different camera views and ShanghaiTech dataset\cite{shanghaitech}, widely utilized resource, provides a total of 295,495 pose instances with thirteen camera views as shown in \cref{tab:dataset}. 

Despite ShanghaiTech's\cite{shanghaitech} diversity, including thirteen distinct scenes, its limited pose instance count per scene constrains its utility for exploration into the feasibility of online anomaly detection across varied domains. In this study, ShanghaiTech\cite{shanghaitech} is trained as the initial source weight for different models and four different camera views from CHAD\cite{danesh2023chad} are used to replicate the stream inputs from four different domains. Notably, Cam 1 to 3 in CHAD\cite{danesh2023chad} have at least 200k poses, providing a substantial volume of data conducive to effective online training and all the train set data are augmented with anomalous pose data extracted from the test set, at a ratio of 9.5:0.5 to mirror the nature of anomalies in typical surveillance scenarios. This strategy ensures that models are exposed to both normal and anomalous patterns.

\subsubsection{Detection}
The pre-processed pose sequences undergo actual inference phase, where they are analyzed within distinct temporal windows size to the architectural requisites of specific models—30 frames for TSGAD\cite{noghre2024exploratory} and GEPC \cite{markovitz2020graph}, and 24 frames for STG-NF\cite{hirschorn2023normalizing} because of specific design. 
This window size is selected to align with the standard frame rate of surveillance cameras. This alignment ensures that the models are not only compatible with standard surveillance video characteristics but also optimized for detecting anomalies within a temporal context that mirrors real-world surveillance scenarios.

\subsection{Collection Methodology}
One primary limitation when utilizing pre-existing datasets for such online anomaly detection design is the constrained volume of pose instances. Moreover, a significant uncertainty in it is the quantity of data that can be classified as "normal" at inference stage with keeping updated weights. To mitigate these challenges and more accurately mirror real-world environments, the training data for each camera view is strategically partitioned into twelve distinct subsets. Once each subset has been inferenced and collected, the training phase would start training. This collecting mechanism allows for the systematic analysis of the models' adaptability and performance across varying conditions, effectively capturing the evolution of domain-specific characteristics.

\subsection{Training Methodology}
In the training phase for each algorithm, default settings were retained with the exception of window and stride sizes. The window size wasadjusted to match the frame rate of the input stream for each specific model, ensuring temporal alignment with the dynamics of the observed activities in one second. The stride size was uniformly set to 1 across all models, such decision aimed at achieving balance within the pipeline's design, particularly to optimize the efficiency and responsiveness of the inference stage.

Leveraging pre-existing datasets for fine-tuning offers the distinct advantage of enabling evaluation of each model's performance with different input subsets using the respective test sets. To thoroughly evaluate the performance of models and gain a multifaceted understanding of their strengths and weaknesses, especially in real-world scenarios, we selected a comprehensive suite of metrics. These metrics—Area Under the Receiver Operating Characteristic Curve (AUC-ROC), Area Under the Precision-Recall Curve (AUC-PR), and Equal Error Rate (EER)—are utilized to assess the efficacy of the models from complementary perspectives, ensuring a holistic analysis.

\textbf{AUC-ROC} is a performance measurement for binary classification problems at various threshold settings. The ROC curve is a graphical representation that plots the True Positive Rate (TPR) against the False Positive Rate (FPR) at different thresholds, essentially showing the trade-off between sensitivity and specificity. The AUC represents the degree to which a model is capable of distinguishing between classes. The higher the AUC, the better the model is at discriminating between the classes. The AUC-ROC metric, while useful in many scenarios, can indeed be misleading in the context of highly imbalanced datasets, such as those typically found in anomaly detection. Thus, it is vital to use it in combination with other metrics to analyze the efficacy of anomaly detection models thoroughly.

\textbf{AUC-PR} is a metric that evaluates the trade-off between precision (the proportion of true positive results in all positive predictions) and recall (the proportion of true positive results in all actual positives) for different threshold values, without being affected by the distribution of class labels. This makes AUC-PR especially valuable for analyzing the efficacy of anomaly detection models, where data is often imbalanced. The AUC-PR encapsulates the model's ability to identify the rare positive cases (anomalies) correctly while minimizing false positives, which is crucial in anomaly detection scenarios where the primary concern is the accurate detection of these rare events.

\textbf{EER} represents the point at which the FPR and False Negative Rate (FNR) are equal. In the context of anomaly detection, the EER offers a singular, balanced threshold at which the likelihood of incorrectly labeling normal behavior as an anomaly equals the likelihood of failing to detect an actual anomaly. The EER aids in identifying the optimal operating point of a model, thereby facilitating more informed decisions in the deployment of anomaly detection systems.

\begin{table}[]
\centering
\caption{Evaluation of Models Pre-trained on Shanghaitech\cite{shanghaitech}}
\label{tab:no train}
\resizebox{\columnwidth}{!}{%
\begin{tabular}{c|c|c|c|c}
\textbf{Model} & \textbf{Test} & \textbf{AUC-ROC} & \textbf{AUC-PR} & \textbf{EER} \\ \hline
\multirow{5}{*}{\textbf{TSGAD}\cite{noghre2024exploratory}} & ShanghaiTech\cite{shanghaitech} & 0.742 & 0.602 & 0.315 \\ \cline{2-5} 
 & CHAD\cite{danesh2023chad} Cam 0 & 0.549 & 0.550 & 0.494 \\ \cline{2-5} 
 & CHAD\cite{danesh2023chad} Cam 1 & 0.561 & 0.487 & 0.467 \\ \cline{2-5} 
 & CHAD\cite{danesh2023chad} Cam 2 & 0.477 & 0.382 & 0.507 \\ \cline{2-5} 
 & CHAD\cite{danesh2023chad} Cam 3 & 0.638 & 0.696 & 0.498 \\ \hline
\multirow{5}{*}{\textbf{GEPC}\cite{markovitz2020graph}} & ShanghaiTech\cite{shanghaitech} & 0.729 & 0.614 & 0.318 \\ \cline{2-5} 
 & CHAD\cite{danesh2023chad} Cam 0 & 0.623 & 0.608 & 0.409 \\ \cline{2-5} 
 & CHAD\cite{danesh2023chad} Cam 1 & 0.622 & 0.491 & 0.407 \\ \cline{2-5} 
 & CHAD\cite{danesh2023chad} Cam 2 & 0.592 & 0.494 & 0.437 \\ \cline{2-5} 
 & CHAD\cite{danesh2023chad} Cam 3 & 0.680 & 0.693 & 0.370 \\ \hline
\multirow{5}{*}{\textbf{STG-NF}\cite{hirschorn2023normalizing}} & ShanghaiTech\cite{shanghaitech} & 0.851 & 0.869 & 0.230 \\ \cline{2-5} 
 & CHAD\cite{danesh2023chad} Cam 0 & 0.582 & 0.638 & 0.459 \\ \cline{2-5} 
 & CHAD\cite{danesh2023chad} Cam 1 & 0.550 & 0.634 & 0.488 \\ \cline{2-5} 
 & CHAD\cite{danesh2023chad} Cam 2 & 0.498 & 0.608 & 0.495 \\ \cline{2-5} 
 & CHAD\cite{danesh2023chad} Cam 3 & 0.633 & 0.573 & 0.430
\end{tabular}%
}
\end{table}

\section{Experiments and Evaluation}
\label{sec:experiments}

As mentioned in \cref{sec:methodology}, ShanghaiTech\cite{shanghaitech} serves as the foundation for initial model training, providing pre-trained weights representative of the source domain.  The efficacy of these initial weights, derived from various models, was evaluated across multiple test sets, including those from ShanghaiTech\cite{shanghaitech} and diverse camera views within the CHAD dataset\cite{danesh2023chad}. The comparative performance analysis is summarized in \cref{tab:no train}

\begin{table*}[]
\centering
\caption{Evaluation of TSGAD\cite{noghre2024exploratory}, GEPC\cite{markovitz2020graph}, and STG-NF\cite{hirschorn2023normalizing} on different camera views from CHAD\cite{danesh2023chad} in cases of Baseline (No Train), Online Training, and Offline Training. }
\label{tab:online}
\begin{tabular}{ccccc|ccccc}
\multicolumn{1}{c|}{\textbf{Model}} & \multicolumn{1}{c|}{\textbf{Case}} & \multicolumn{1}{c|}{\textbf{AUC-ROC}} & \multicolumn{1}{c|}{\textbf{AUC-PR}} & \textbf{EER} & \multicolumn{1}{c|}{\textbf{Model}} & \multicolumn{1}{c|}{\textbf{Case}} & \multicolumn{1}{c|}{\textbf{AUC-ROC}} & \multicolumn{1}{c|}{\textbf{AUC-PR}} & \textbf{EER} \\ \hline
\multicolumn{5}{c|}{\textbf{Cam 0}} & \multicolumn{5}{c}{\textbf{Cam 1}} \\ \hline
\multicolumn{1}{c|}{\multirow{3}{*}{\textbf{TSGAD}\cite{noghre2024exploratory}}} & \multicolumn{1}{c|}{No Train} & \multicolumn{1}{c|}{0.549} & \multicolumn{1}{c|}{0.550} & 0.494 & \multicolumn{1}{c|}{\multirow{3}{*}{\textbf{TSGAD}\cite{noghre2024exploratory}}} & \multicolumn{1}{c|}{No Train} & \multicolumn{1}{c|}{0.561} & \multicolumn{1}{c|}{0.487} & 0.467 \\ \cline{2-5} \cline{7-10} 
\multicolumn{1}{c|}{} & \multicolumn{1}{c|}{Online} & \multicolumn{1}{c|}{0.565} & \multicolumn{1}{c|}{0.548} & 0.466 & \multicolumn{1}{c|}{} & \multicolumn{1}{c|}{Online} & \multicolumn{1}{c|}{0.568} & \multicolumn{1}{c|}{0.488} & 0.461 \\ \cline{2-5} \cline{7-10} 
\multicolumn{1}{c|}{} & \multicolumn{1}{c|}{Offline} & \multicolumn{1}{c|}{0.632} & \multicolumn{1}{c|}{0.608} & 0.409 & \multicolumn{1}{c|}{} & \multicolumn{1}{c|}{Offline} & \multicolumn{1}{c|}{0.601} & \multicolumn{1}{c|}{0.506} & 0.430 \\ \hline
\multicolumn{1}{c|}{\multirow{3}{*}{\textbf{GEPC}\cite{markovitz2020graph}}} & \multicolumn{1}{c|}{No Train} & \multicolumn{1}{c|}{0.623} & \multicolumn{1}{c|}{0.608} & 0.409 & \multicolumn{1}{c|}{\multirow{3}{*}{\textbf{GEPC}\cite{markovitz2020graph}}} & \multicolumn{1}{c|}{No Train} & \multicolumn{1}{c|}{0.622} & \multicolumn{1}{c|}{0.491} & 0.407 \\ \cline{2-5} \cline{7-10} 
\multicolumn{1}{c|}{} & \multicolumn{1}{c|}{Online} & \multicolumn{1}{c|}{0.625} & \multicolumn{1}{c|}{0.625} & 0.419 & \multicolumn{1}{c|}{} & \multicolumn{1}{c|}{Online} & \multicolumn{1}{c|}{0.609} & \multicolumn{1}{c|}{0.495} & 0.422 \\ \cline{2-5} \cline{7-10} 
\multicolumn{1}{c|}{} & \multicolumn{1}{c|}{Offline} & \multicolumn{1}{c|}{0.630} & \multicolumn{1}{c|}{0.635} & 0.415 & \multicolumn{1}{c|}{} & \multicolumn{1}{c|}{Offline} & \multicolumn{1}{c|}{0.630} & \multicolumn{1}{c|}{0.494} & 0.383 \\ \hline
\multicolumn{1}{c|}{\multirow{3}{*}{\textbf{STG-NF}\cite{hirschorn2023normalizing}}} & \multicolumn{1}{c|}{No Train} & \multicolumn{1}{c|}{0.582} & \multicolumn{1}{c|}{0.638} & 0.459 & \multicolumn{1}{c|}{\multirow{3}{*}{\textbf{STG-NF}\cite{hirschorn2023normalizing}}} & \multicolumn{1}{c|}{No Train} & \multicolumn{1}{c|}{0.550} & \multicolumn{1}{c|}{0.634} & 0.488 \\ \cline{2-5} \cline{7-10} 
\multicolumn{1}{c|}{} & \multicolumn{1}{c|}{Online} & \multicolumn{1}{c|}{0.596} & \multicolumn{1}{c|}{0.651} & 0.442 & \multicolumn{1}{c|}{} & \multicolumn{1}{c|}{Online} & \multicolumn{1}{c|}{0.574} & \multicolumn{1}{c|}{0.661} & 0.456 \\ \cline{2-5} \cline{7-10} 
\multicolumn{1}{c|}{} & \multicolumn{1}{c|}{Offline} & \multicolumn{1}{c|}{0.615} & \multicolumn{1}{c|}{0.667} & 0.429 & \multicolumn{1}{c|}{} & \multicolumn{1}{c|}{Offline} & \multicolumn{1}{c|}{0.562} & \multicolumn{1}{c|}{0.654} & 0.464 \\ \hline
\multicolumn{5}{c|}{\textbf{Cam 3}} & \multicolumn{5}{c}{\textbf{Cam 4}} \\ \hline
\multicolumn{1}{c|}{\multirow{3}{*}{\textbf{TSGAD}\cite{noghre2024exploratory}}} & \multicolumn{1}{c|}{No Train} & \multicolumn{1}{c|}{0.477} & \multicolumn{1}{c|}{0.382} & 0.507 & \multicolumn{1}{c|}{\multirow{3}{*}{\textbf{TSGAD}\cite{noghre2024exploratory}}} & \multicolumn{1}{c|}{No Train} & \multicolumn{1}{c|}{0.638} & \multicolumn{1}{c|}{0.696} & 0.498 \\ \cline{2-5} \cline{7-10} 
\multicolumn{1}{c|}{} & \multicolumn{1}{c|}{Online} & \multicolumn{1}{c|}{0.497} & \multicolumn{1}{c|}{0.393} & 0.493 & \multicolumn{1}{c|}{} & \multicolumn{1}{c|}{Online} & \multicolumn{1}{c|}{0.647} & \multicolumn{1}{c|}{0.706} & 0.397 \\ \cline{2-5} \cline{7-10} 
\multicolumn{1}{c|}{} & \multicolumn{1}{c|}{Offline} & \multicolumn{1}{c|}{0.561} & \multicolumn{1}{c|}{0.490} & 0.422 & \multicolumn{1}{c|}{} & \multicolumn{1}{c|}{Offline} & \multicolumn{1}{c|}{0.655} & \multicolumn{1}{c|}{0.656} & 0.385 \\ \hline
\multicolumn{1}{c|}{\multirow{3}{*}{\textbf{GEPC}\cite{markovitz2020graph}}} & \multicolumn{1}{c|}{No Train} & \multicolumn{1}{c|}{0.592} & \multicolumn{1}{c|}{0.494} & 0.437 & \multicolumn{1}{c|}{\multirow{3}{*}{\textbf{GEPC}\cite{markovitz2020graph}}} & \multicolumn{1}{c|}{No Train} & \multicolumn{1}{c|}{0.680} & \multicolumn{1}{c|}{0.693} & 0.370 \\ \cline{2-5} \cline{7-10} 
\multicolumn{1}{c|}{} & \multicolumn{1}{c|}{Online} & \multicolumn{1}{c|}{0.596} & \multicolumn{1}{c|}{0.505} & 0.429 & \multicolumn{1}{c|}{} & \multicolumn{1}{c|}{Online} & \multicolumn{1}{c|}{0.685} & \multicolumn{1}{c|}{0.704} & 0.350 \\ \cline{2-5} \cline{7-10} 
\multicolumn{1}{c|}{} & \multicolumn{1}{c|}{Offline} & \multicolumn{1}{c|}{0.625} & \multicolumn{1}{c|}{0.510} & 0.391 & \multicolumn{1}{c|}{} & \multicolumn{1}{c|}{Offline} & \multicolumn{1}{c|}{0.693} & \multicolumn{1}{c|}{0.695} & 0.338 \\ \hline
\multicolumn{1}{c|}{\multirow{3}{*}{\textbf{STG-NF}\cite{hirschorn2023normalizing}}} & \multicolumn{1}{c|}{No Train} & \multicolumn{1}{c|}{0.498} & \multicolumn{1}{c|}{0.608} & 0.495 & \multicolumn{1}{c|}{\multirow{3}{*}{\textbf{STG-NF}\cite{hirschorn2023normalizing}}} & \multicolumn{1}{c|}{No Train} & \multicolumn{1}{c|}{0.633} & \multicolumn{1}{c|}{0.573} & 0.423 \\ \cline{2-5} \cline{7-10} 
\multicolumn{1}{c|}{} & \multicolumn{1}{c|}{Online} & \multicolumn{1}{c|}{0.510} & \multicolumn{1}{c|}{0.621} & 0.495 & \multicolumn{1}{c|}{} & \multicolumn{1}{c|}{Online} & \multicolumn{1}{c|}{0.647} & \multicolumn{1}{c|}{0.585} & 0.399 \\ \cline{2-5} \cline{7-10} 
\multicolumn{1}{c|}{} & \multicolumn{1}{c|}{Offline} & \multicolumn{1}{c|}{0.520} & \multicolumn{1}{c|}{0.626} & 0.495 & \multicolumn{1}{c|}{} & \multicolumn{1}{c|}{Offline} & \multicolumn{1}{c|}{0.659} & \multicolumn{1}{c|}{0.600} & 0.392
\end{tabular}%
\end{table*}

Notably, STG-NF\cite{hirschorn2023normalizing} emerged as the best model within the ShanghaiTech\cite{shanghaitech} test set, outperforming others across all evaluated metrics, thereby affirming its status as a state-of-the-art (SoTA) algorithm. TSGAD(pose only)\cite{noghre2024exploratory}, which originally has pose and path branch, ranked second, showing its strength in AUC-ROC and EER metrics. However, when subjected to test sets from different domains, both STG-NF\cite{hirschorn2023normalizing} and TSGAD(pose only)\cite{noghre2024exploratory} experienced significant declines in accuracy, underscoring the challenge of domain shift. In contrast, GEPC\cite{markovitz2020graph} shows relatively stable performance across varied test environments, achieving the highest overall scores in all CHAD\cite{danesh2023chad} camera views. Following with TSGAD(pose only)\cite{noghre2024exploratory} performs the second in CHAD\cite{danesh2023chad} CAM 1 and CAM 3 and third in CAM 0 and CAM 2. 

This pattern of results, notable drop in accuracy in different domain, aligns with the expectation highlighted in \cref{sec:methodology} that anomaly detection algorithms are highly sensitive to contextual variations.

\begin{figure*}[htbp]
  \centering
  \begin{subfigure}[b]{0.43\textwidth}
    \includegraphics[width=\textwidth, trim= 12 23 15 74, clip]{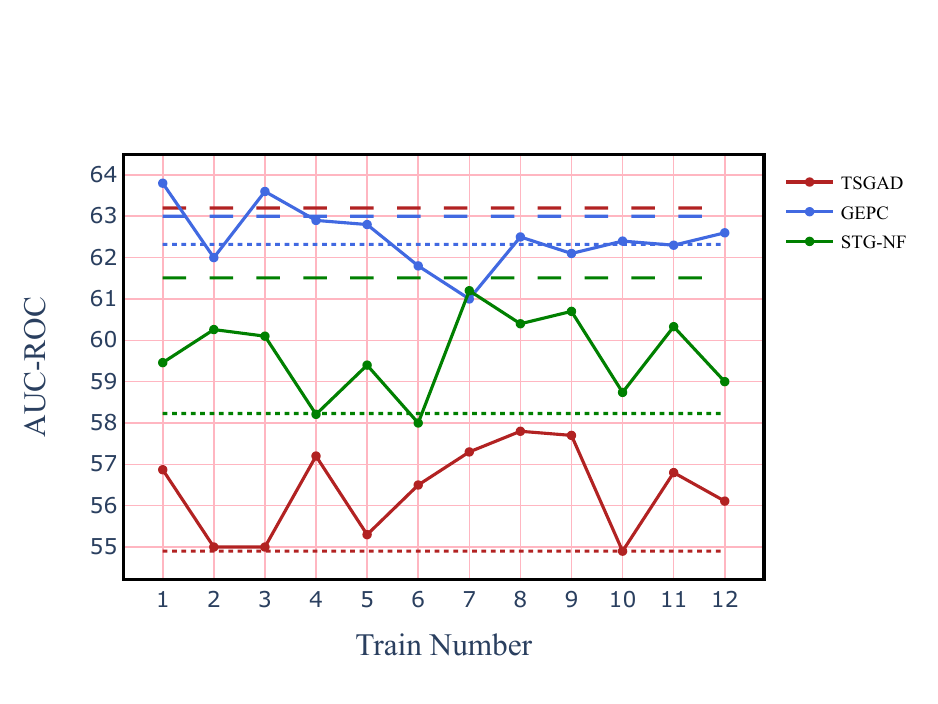}
    \caption{Cam 0}
    \label{fig:1}
  \end{subfigure}
  \hfill
  \begin{subfigure}[b]{0.43\textwidth}
    \includegraphics[width=\textwidth, trim= 12 23 15 74, clip]{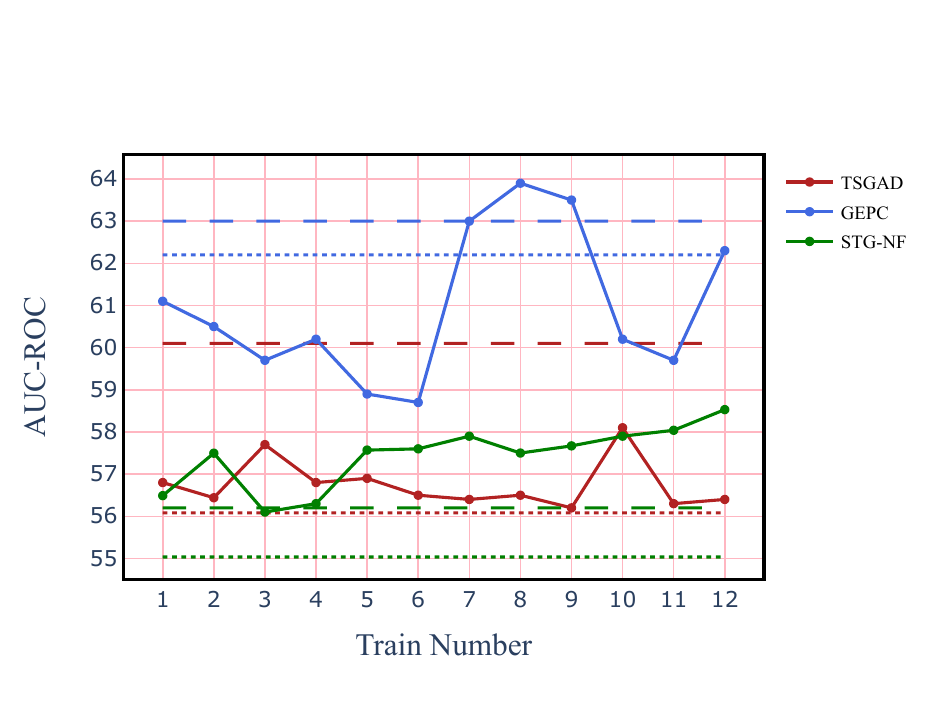}
    \caption{Cam 1}
    \label{fig:2}
  \end{subfigure}
  
  \medskip
  
  \begin{subfigure}[b]{0.43\textwidth}
    \includegraphics[width=\textwidth, trim= 12 23 15 74, clip]{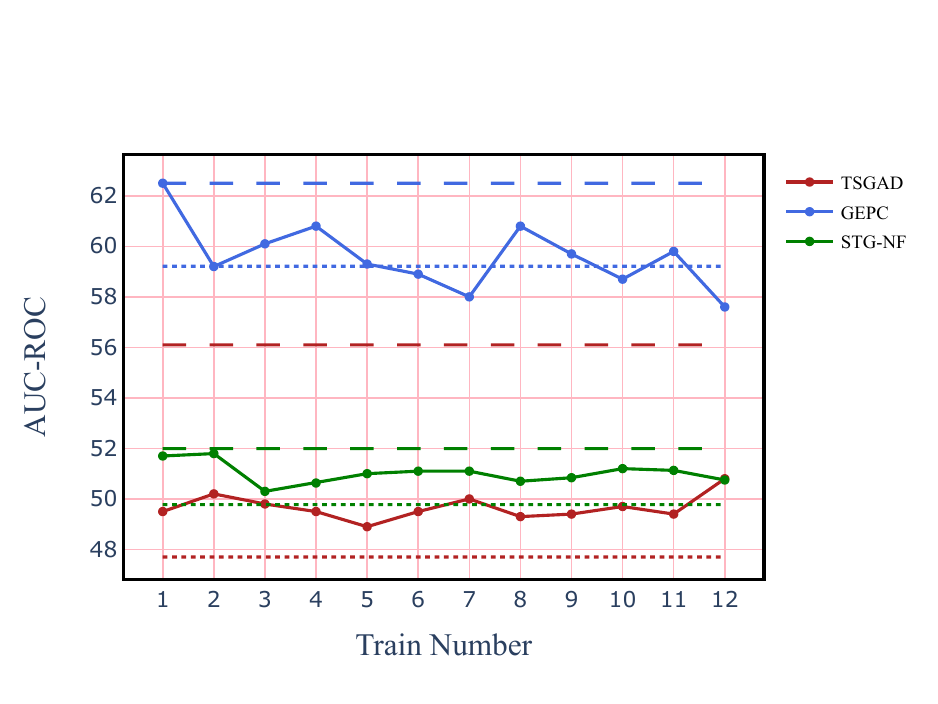}
    \caption{Cam 2}
    \label{fig:3}
  \end{subfigure}
  \hfill
  \begin{subfigure}[b]{0.43\textwidth}
    \includegraphics[width=\textwidth, trim= 12 23 15 74, clip]{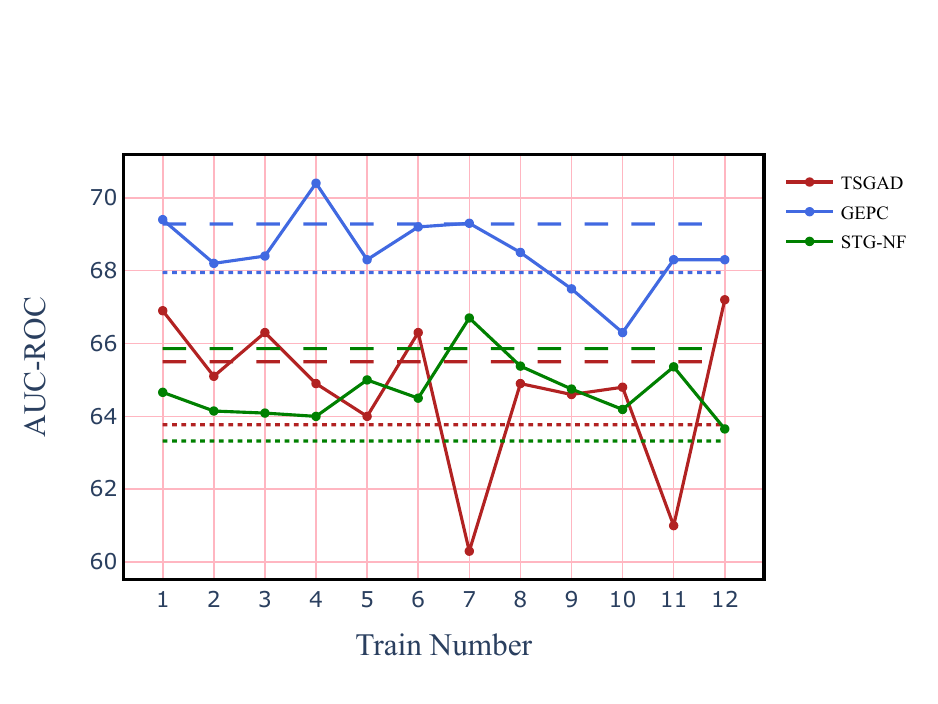}
    \caption{Cam 3}
    \label{fig:4}
  \end{subfigure}
  \caption{Model AUC-ROC percentage Trend Comparison by Training Number: Long dashes indicate Offline Training, solid lines indicate Online Training, and dots indicate the Baseline (No training).}
  \label{fig:roc}
\end{figure*}

\begin{figure*}[htbp]
  \centering
  \begin{subfigure}[b]{0.43\textwidth}
    \includegraphics[width=\textwidth, trim= 12 23 15 74, clip]{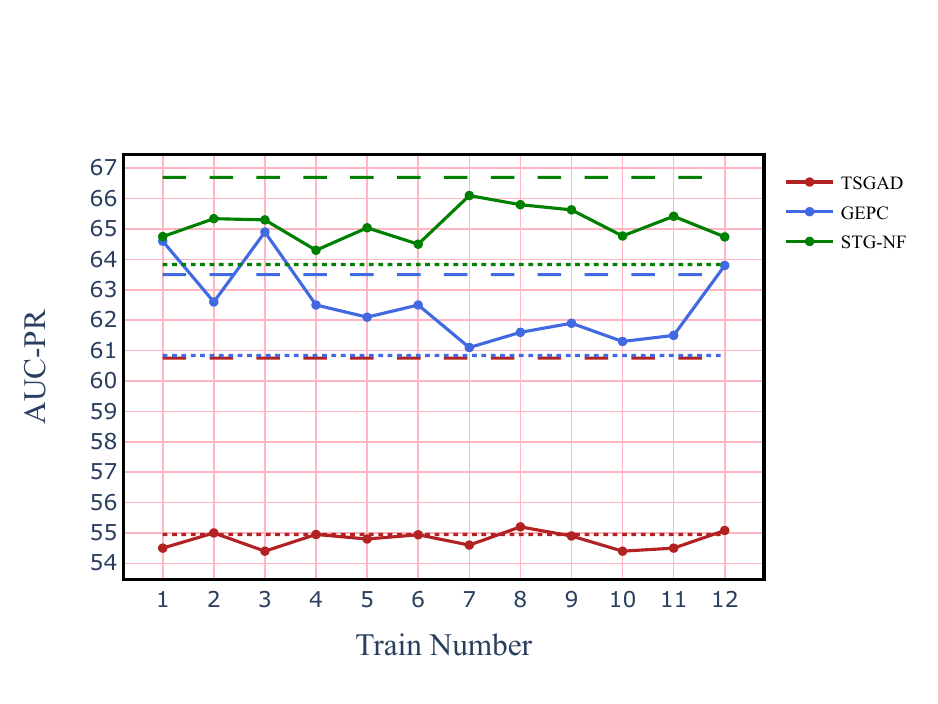}
    \caption{Cam 0}
    \label{fig:1}
  \end{subfigure}
  \hfill
  \begin{subfigure}[b]{0.43\textwidth}
    \includegraphics[width=\textwidth, trim= 12 23 15 74, clip]{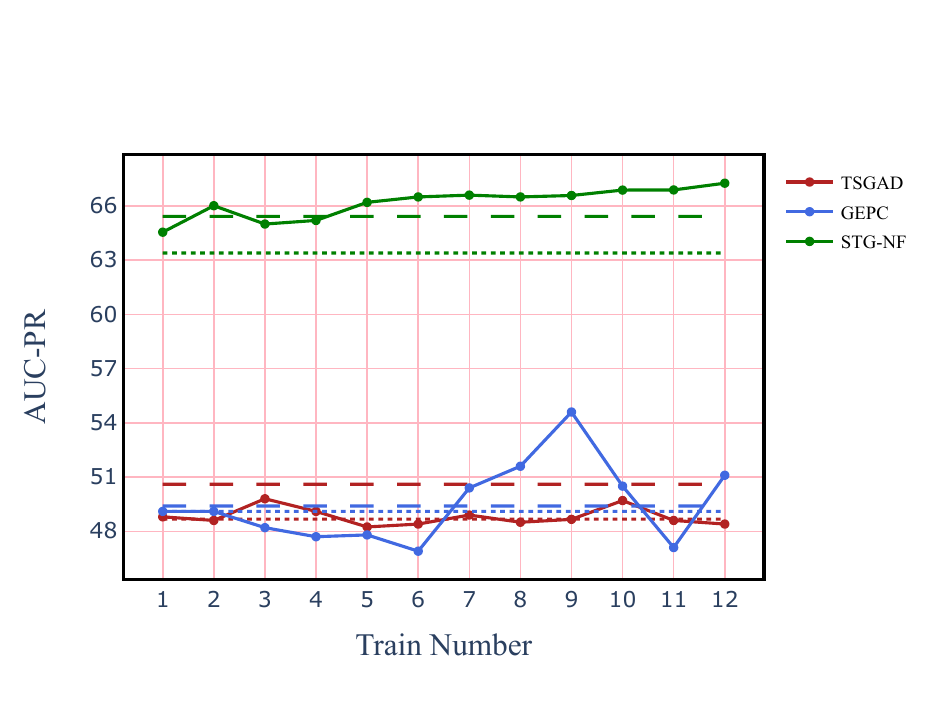}
    \caption{Cam 1}
    \label{fig:2}
  \end{subfigure}
  
  \medskip
  
  \begin{subfigure}[b]{0.43\textwidth}
    \includegraphics[width=\textwidth, trim= 12 23 15 74, clip]{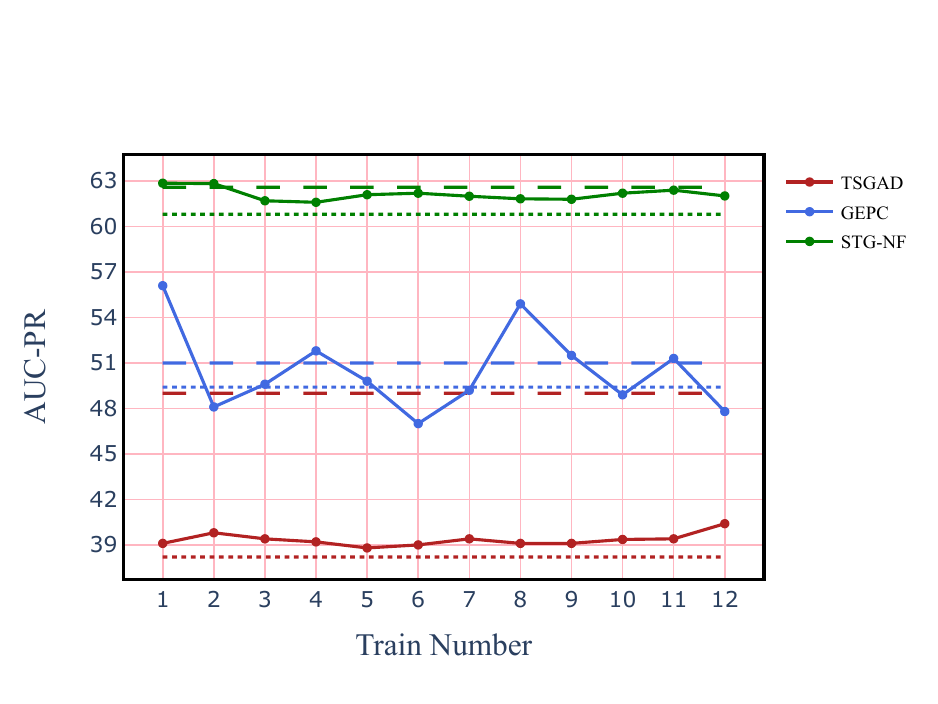}
    \caption{Cam 2}
    \label{fig:3}
  \end{subfigure}
  \hfill
  \begin{subfigure}[b]{0.43\textwidth}
    \includegraphics[width=\textwidth, trim= 12 23 15 74, clip]{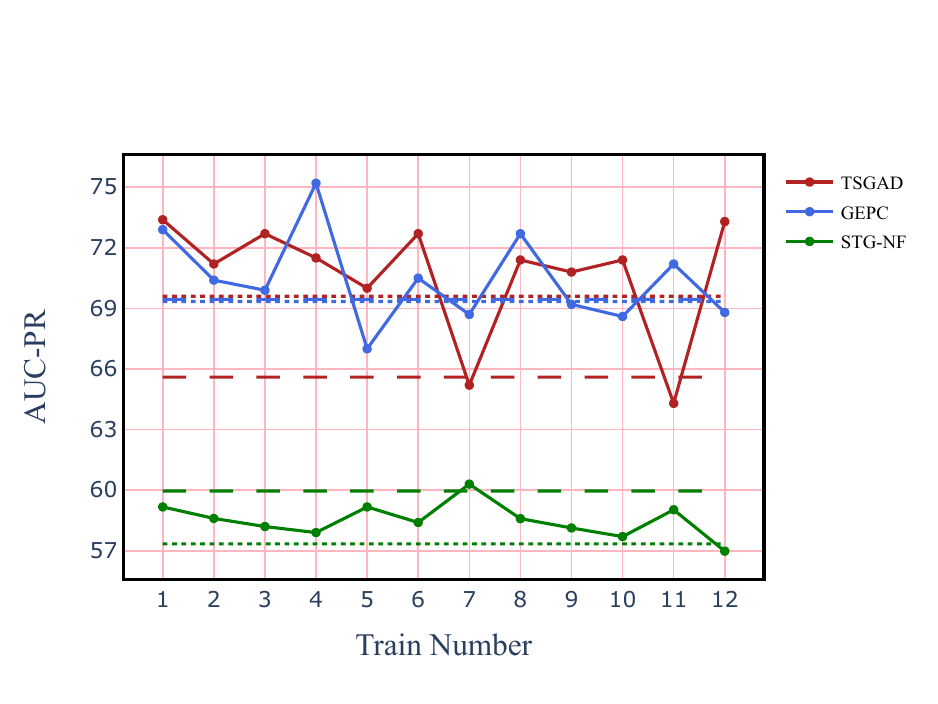}
    \caption{Cam 3}
    \label{fig:4}
  \end{subfigure}
  \caption{Model AUC-PR percentage Trend Comparison by Training Number: Long dashes indicate Offline Training, solid lines indicate Online Training, and dots indicate the Baseline (No training).}
  \label{fig:pr}
\end{figure*}

\begin{figure*}[htbp]
  \centering
  \begin{subfigure}[b]{0.43\textwidth}
    \includegraphics[width=\textwidth, trim= 12 23 15 74, clip]{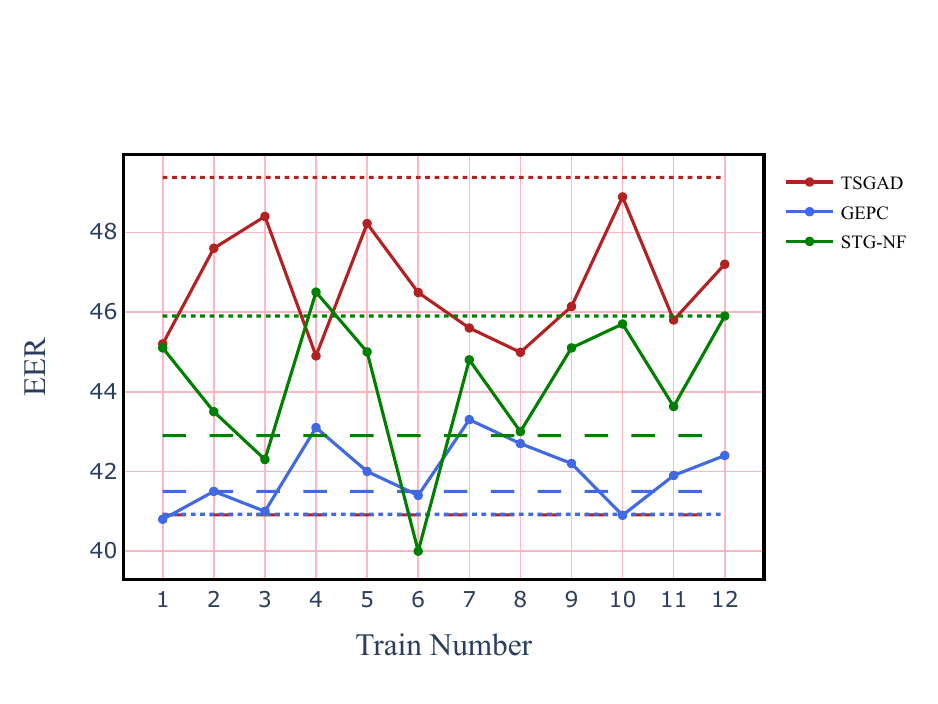}
    \caption{Cam 0}
    \label{fig:1}
  \end{subfigure}
  \hfill
  \begin{subfigure}[b]{0.43\textwidth}
    \includegraphics[width=\textwidth, trim= 12 23 15 74, clip]{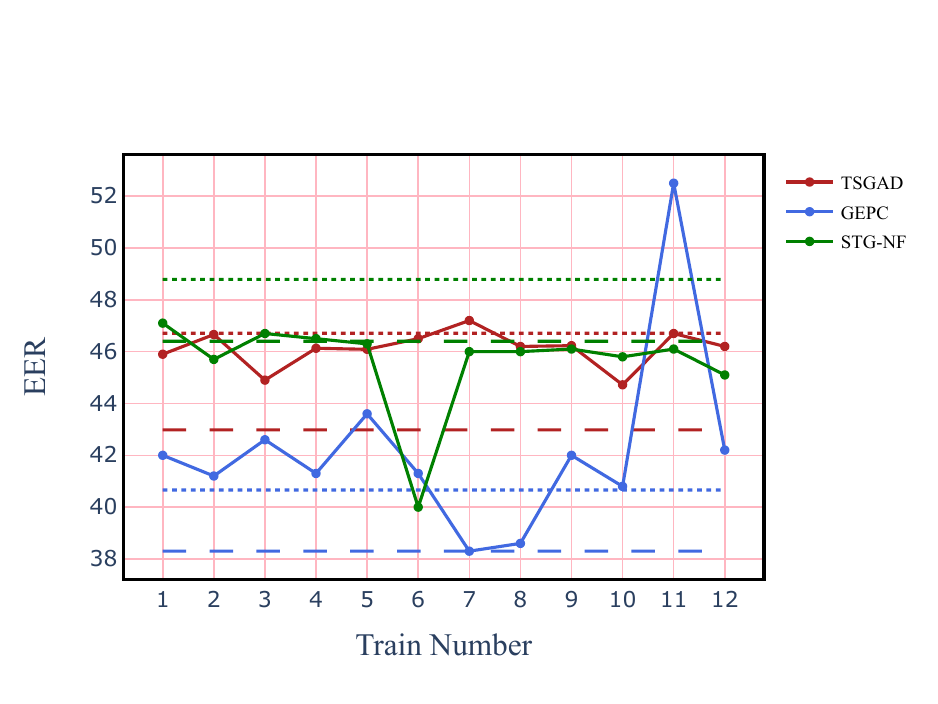}
    \caption{Cam 1}
    \label{fig:2}
  \end{subfigure}
  
  \medskip
  
  \begin{subfigure}[b]{0.43\textwidth}
    \includegraphics[width=\textwidth, trim= 12 23 15 74, clip]{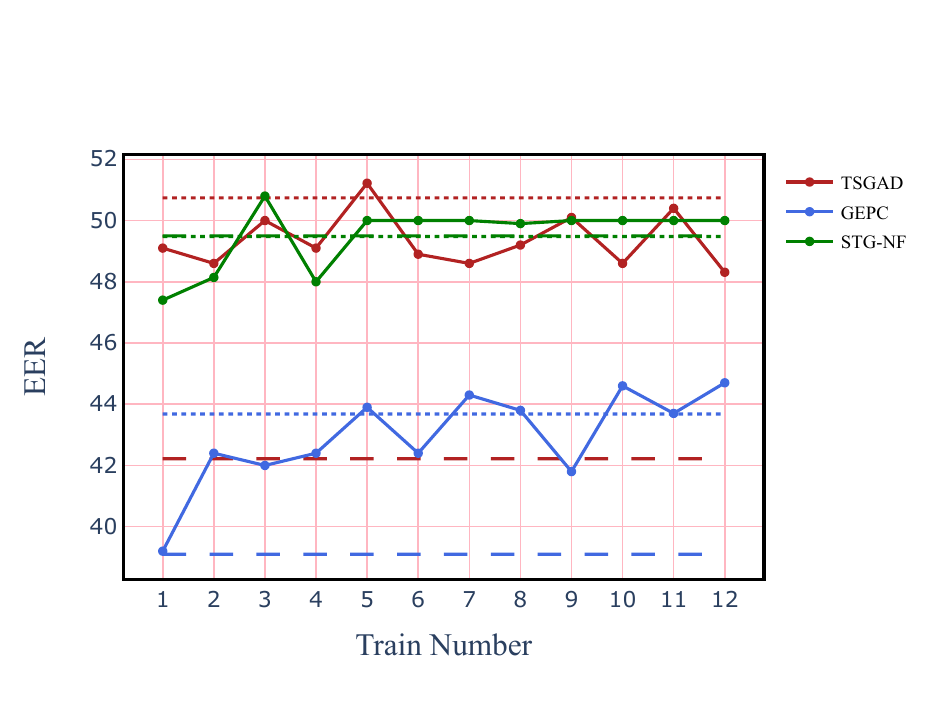}
    \caption{Cam 2}
    \label{fig:3}
  \end{subfigure}
  \hfill
  \begin{subfigure}[b]{0.43\textwidth}
    \includegraphics[width=\textwidth, trim= 12 23 15 74, clip]{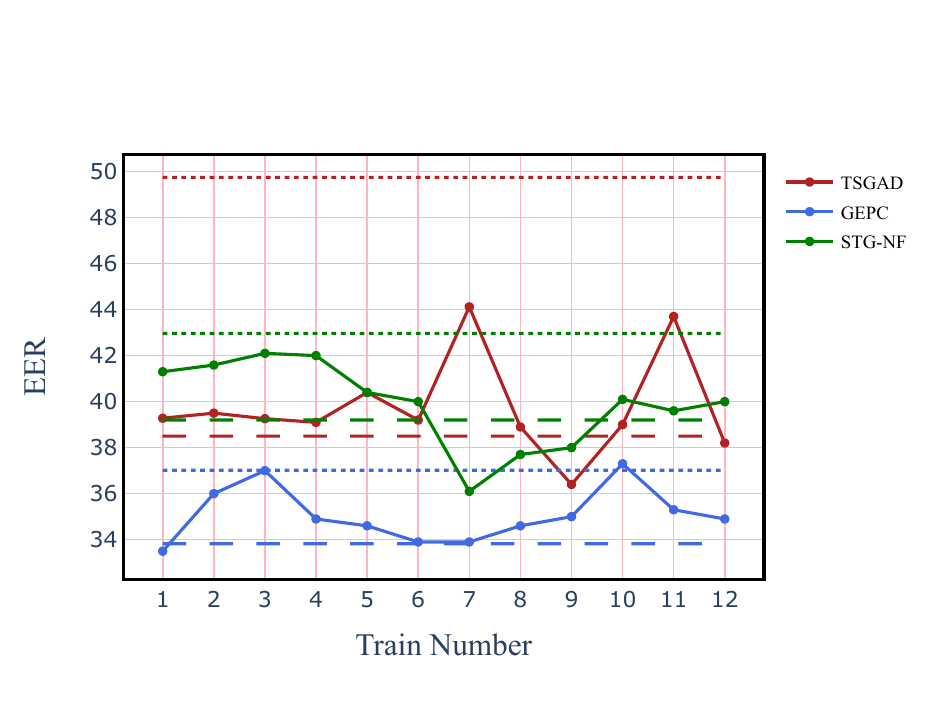}
    \caption{Cam 3}
    \label{fig:4}
  \end{subfigure}
  \caption{Model EER percentage Trend Comparison by Training Number: Long dashes indicate Offline Training, solid lines indicate Online Training, and dots indicate the Baseline (No training).}
  \label{fig:eer}
\end{figure*}

\cref{tab:online} presents a comparison of model evaluation result across three distinct training scenarios: baseline (no training), average performance across twelve online training, and outcomes following complete offline training. The evaluation spans four domains (Cam 0 to Cam 3), each offering unique insights into the adaptability and efficacy of the models under consideration: TSGAD(pose only)\cite{noghre2024exploratory}. GEPC\cite{markovitz2020graph}, and STG-NF\cite{hirschorn2023normalizing}. 

\textbf{Cam 0} In the online training scenario, GEPC\cite{markovitz2020graph} emerged as the top performer, with STG-NF\cite{hirschorn2023normalizing} closely following, showing their adaptability to this specific domain. Conversely, the offline training scenario highlighted TSGAD's\cite{noghre2024exploratory} fine performance, with notable improvements observed from the baseline to the online training phase, indicating TSGAD's\cite{noghre2024exploratory} learning capability in this domain.

\textbf{Cam 1} GEPC\cite{markovitz2020graph} consistently achieved good results across most evaluation cases, yet an unexpected drop in performance was observed transitioning from the no training to the online training scenario. This suggests a potential anomaly in GEPC's\cite{markovitz2020graph} learning curve or an overfitting issue within this specific context. STG-NF\cite{hirschorn2023normalizing}, while generally underperforming, exhibited a further decline in the offline training scenario, raising questions about its adaptability and efficacy.

\textbf{Cam 2} Despite a general decline in model performance in this domain, GEPC\cite{markovitz2020graph} maintained its lead, followed by TSGAD(pose only)\cite{noghre2024exploratory} and STG-NF\cite{hirschorn2023normalizing}. This consistent pattern across models suggests inherent challenges within the Cam 2 domain.

\textbf{Cam 3} TSGAD(pose only)\cite{noghre2024exploratory} and STG-NF\cite{hirschorn2023normalizing} demonstrated remarkably similar performance metrics, indicating a convergence in their learning capabilities for this camera view. GEPC\cite{markovitz2020graph}, maintaining its pattern, stood out with its performance, reinforcing its robustness across varied conditions.

Detailed analyses of the models' performance are visualized in \cref{fig:roc},
\cref{fig:pr}, and \cref{fig:eer}, each illustrating variations in evaluation metrics across training numbers. GEPC\cite{markovitz2020graph} consistently exhibits strong performance in AUC-ROC and EER metrics, while STG-NF\cite{hirschorn2023normalizing} excels in AUC-PR across multiple training stages. Notably,  STG-NF\cite{hirschorn2023normalizing} model consistently shows high EER and low AUC-ROC, yet achieving high AUC-PR. Despite this, STG-NF\cite{hirschorn2023normalizing} could be useful in situations where minimizing false negatives is more important than avoiding false positives. The model's tendency to mistakenly classify normal instances as anomalies, leading to high EER and low AUC-ROC, may be due to its simplistic assumption of a normal distribution in the latent space, which struggles to capture complex patterns in large datasets like CHAD\cite{danesh2023chad}.

While GEPC\cite{markovitz2020graph} generally exhibited superior performance across multiple evaluation scenarios, its incremental learning gains from baseline to complete offline training were modest, rarely exceeding a 2\% improvement. This phenomenon likely caused by the design of GEPC's\cite{markovitz2020graph} algorithm, which incorporates a sequencial training mechanism with encoding, decoding, and clustering phases. The necessity to start each training cycle with encoder fine-tuning, effectively initializing decoder and cluster components without pre-trained weights, might be not ideal for the demands of online anomaly training. This design choice, while beneficial in certain contexts, appears to constrain GEPC's\cite{markovitz2020graph} capacity in online anomaly training design.

On the other hand, STG-NF\cite{hirschorn2023normalizing} demonstrated incredible performance within the controlled setting of the ShanghaiTech dataset\cite{shanghaitech} during offline training. However, its adaptability to the varied domains represented in the CHAD dataset\cite{danesh2023chad} was less consistent, with improvements from the baseline to offline training being modest, at approximately 4\%. This suggests that while STG-NF\cite{hirschorn2023normalizing} is capable of learning and improving, its architecture may not be suited to the diverse and dynamic nature of real-world surveillance scenarios in online anomaly training.

TSGAD(pose only)\cite{noghre2024exploratory}, although not consistently surpassing GEPC\cite{markovitz2020graph} in domain-specific evaluations, exhibited notable progressions from baseline through online training to offline training. This trajectory of improvement highlights TSGAD's\cite{noghre2024exploratory} potential compatibility with online anomaly training frameworks. The model's ability to learn effectively suggests a structural or algorithmic adaptability that could be optimized for the continuous, evolving nature of online anomaly training. 

These observations underscore the nuanced relationship between model architecture, training methodology, and domain specificity in anomaly detection. The varied performance and learning trajectories of GEPC\cite{markovitz2020graph}, STG-NF\cite{hirschorn2023normalizing}, and TSGAD(pose only)\cite{noghre2024exploratory} across different training stages and domains illuminate the critical considerations necessary for tailoring anomaly detection models to the specific requirements and challenges of online training environments.

\section{Research Questions and Future Directions}
\label{sec:future}
This study highlights the viability and potential of employing online training strategy with pose-based anomaly detection models utilizing existing datasets. As mentioned in \cref{sec:methodology}, the reliance on current datasets introduces uncertainties related to the volume of training data each training time, complicating the exploration of time constraints within online training frameworks.

It highlights the challenges posed by variable training data volumes and underscores the necessity of an integrated end-to-end system for live stream processing and training execution in future research. Despite these challenges, our method retains 89.38\% effectiveness relative to offline training, suggesting the potential to not only match but exceed offline training efficacy with further enhancements. 
Critical to this ambition is the precise measurement of the online learning timing in a balanced system, essential for demonstrating the practicality of such systems in the wild applications. The objective is to identify and surpass the limitations of existing approaches, thereby designing more effective strategies to tackle the domain-specific barrier encountered in the detection of anomalous actions, thereby advancing the state of research in this specialized area.
 
\section{Conclusion}
\label{sec:conclusion}

This study underscores the significant challenges inherent in applying Video Anomaly Detection (VAD) in real-world scenarios,  due to the dynamic and unpredictable nature of human behavior, environmental contexts, and domain shifts. Through an evaluation of SOTA VAD algorithms within an online learning framework, our research highlights the potential of pose-based approaches to not only address these challenges but also to offer privacy-conscious solutions suitable for practical applications. The adaptability of these models to continuously learn and update from streaming data represents a critical step forward in bridging the gap between theoretical research and real-world applicability. Our findings demonstrate that even under the most challenging conditions, the proposed online learning method enables models to maintain a high degree of effectiveness, retaining up to 89.39\% of their original performance. This work paves the way for future research in enhancing the robustness and adaptability of VAD systems, ensuring their reliability and efficacy in the wild environments.
\section*{Acknowledgment}
This research is supported by the National Science Foundation (NSF) under Award No. 1831795.
{
    \small
    \bibliographystyle{ieeenat_fullname}
    \bibliography{main}
}


\end{document}